\documentclass[11pt]{article}

\usepackage{booktabs}
\usepackage[]{acl}

\usepackage{times}
\usepackage{latexsym}
\usepackage{tabularx}
\usepackage{booktabs}
\usepackage{enumitem}
\usepackage{array}
\usepackage{makecell}
\usepackage{multirow}
\usepackage{amsmath}

\DeclareMathOperator*{\argmin}{arg\,min}
\usepackage{hyperref}
\usepackage{pgfplots}
\usepackage{filecontents}

\newcolumntype{H}{>{\setbox0=\hbox\bgroup}c<{\egroup}@{}}
\usepackage[T1]{fontenc}
\usepackage{xcolor,colortbl}

\definecolor{Gray}{gray}{0.85}

\usepackage[utf8]{inputenc}
\usepackage{tikz,siunitx}
\usetikzlibrary{shapes.geometric,shapes.symbols}
\usepackage{microtype}

\newenvironment{itemizesquish}{\begin{list}{\labelitemi}{\setlength{\itemsep}{0em}\setlength{\labelwidth}{0.5em}\setlength{\leftmargin}{\labelwidth}\addtolength{\leftmargin}{\labelsep}}}{\end{list}}

\title{On ``Scientific Debt'' in NLP: A Case for More Rigour \\ in Language Model Pre-Training Research}

\author{
Made Nindyatama Nityasya$^1$, Haryo Akbarianto Wibowo$^1$, Alham Fikri Aji$^2$, \\ 
\textbf{Genta Indra Winata$^3$, Radityo Eko Prasojo$^4$, Phil Blunsom$^{5,6}$, Adhiguna Kuncoro$^7$} \\
$^1$Independent Researcher \quad $^2$MBZUAI \quad $^3$Bloomberg \quad $^4$Universitas Indonesia \\
$^5$Cohere.AI \quad $^6$University of Oxford \quad $^7$DeepMind\\
  \texttt{\small \{made.nindyatama,haryo.akbarianto\}@gmail.com, alham.fikri@mbzuai.ac.ae,} \\
  \texttt{\small gwinata@bloomberg.net, radityo.ep@ui.ac.id, phil.blunsom@cs.ox.ac.uk, } \\
  \texttt{\small akuncoro@deepmind.com}}

\begin{document}
\maketitle
\begin{abstract}

This evidence-based position paper critiques current research practices within the language model pre-training literature. Despite rapid recent progress afforded by increasingly better pre-trained language models (PLMs), current PLM research practices often conflate different possible sources of model improvement, without conducting proper ablation studies and principled comparisons between different models under comparable conditions. These practices (i) leave us ill-equipped to understand which pre-training approaches should be used under what circumstances; (ii) impede reproducibility and credit assignment; and (iii) render it difficult to understand: ``\emph{How exactly does each factor contribute to the progress that we have today?}'' We provide a case in point by revisiting the success of BERT over its baselines, ELMo and GPT-1, and demonstrate how --- under comparable conditions where the baselines are tuned to a similar extent --- these baselines (and even-simpler variants thereof) can, in fact, achieve competitive or better performance than BERT. These findings demonstrate how disentangling different factors of model improvements can lead to valuable new insights. We conclude with recommendations for how to encourage and incentivize this line of work, and accelerate progress towards a better and more systematic understanding of what factors drive the progress of our foundation models today.

\end{abstract}

\section{Introduction}
In recent years, language models that are pre-trained on large amounts of data have become the foundation models \citep{foundation_models} for achieving state-of-the-art results on many NLP tasks \citep[\emph{inter alia}]{peters-etal-2018-deep,devlin2019bert,raffel2020exploring,liu2019roberta}, and performed well at novel tasks through demonstrations \citep{brown2020language}. Hence, given the vast potential of pre-trained language models (PLMs), substantial effort has since been dedicated into developing the next state-of-the-art PLM, whether through new objective functions \citep[\emph{inter alia}]{devlin2019bert,xlnet,clark2020electra,raffel2020exploring,tay2022unifying}, increasing model size \citep{kaplan2020scaling,brown2020language,chowdhery2022palm}, better data filtering \citep{lee-etal-2022-deduplicating}, or making a better use of the compute budget by training longer \citep{hoffmann2022training}.

To a large extent, this rapid progress is made possible by a strong emphasis on designing better foundation models and achieving new state-of-the-art results. In pursuit of this goal, each new PLM often leverages all possible means for improving performance: Indeed, it is often the case that a new state-of-the-art PLM paper not only proposed a new pre-training loss function as its key novelty, but was also trained on more data, used a larger model size, and benefited from the latest proven hyper-parameter settings and tricks of the trade compared to earlier PLMs. We argue, however, that progress through these practices also comes at a cost: As each new PLM differs from earlier one in \emph{multiple} dimensions at once, it has become increasingly harder to (i) disentangle how these different components contribute to the model performance and progress that we observe today; (ii) understand which approaches work well under what circumstances; (iii) distill generalizable patterns and understand how well each result would transfer to new tasks, datasets, and settings (\emph{e.g.,} in low-resource data and compute scenarios); and (iv) replicate prior results and conduct the appropriate credit assignment for the techniques and prior work that are most responsible for our progress today. 

Much like how \emph{technical debt} often arises when developing new software at breakneck speed,\footnote{Technical debt refers to the results when technical teams develop new software at great speed, which later need to be reworked due to choosing limited solutions at the expense of more generalizable and principled approaches that take longer.} we propose the term ``\textbf{scientific debt}'' to refer to the issues above that arise due to these PLM research practices. In this evidence-based position paper, we argue that scientific progress in NLP should strike a delicate balance between achieving the best performance on various benchmarks and leaderboards --- an area where great progress has been made in recent years --- and also on understanding: \emph{How exactly does each different component affect the PLM performance that we observe today?} While this question is difficult to answer in light of the current PLM research practices that conflate different sources of model performance, we encourage the community to dedicate more effort into disentangling the performance gains from these interacting factors. Doing so would pave the way for achieving more progress in the future, in a way that is more scientifically rigorous, generalizable, reproducible, and \emph{well-grounded} in a better understanding of how well each approach works under different settings. 

We begin by motivating the importance of disentangling multiple possible factors of model improvement through an analogy with medicine, and discuss their 
parallels for PLM research (\S\ref{sec:medicine}). We then provide empirical evidence by revisiting the success of BERT \citep{devlin2019bert}, and demonstrate that prior PLMs  like ELMo \citep{peters-etal-2018-deep} and GPT-1 \citep{radford2018improving} can, in fact, achieve nearly the same performance ($\sim 1\%$ difference in aggregate GLUE) as BERT under \emph{comparable} experimental conditions (\S\ref{sec:empirical}). These findings serve to (i) further our understanding of the effectiveness of the masked language modelling loss compared to prior approaches, under comparable experimental conditions; and (ii) provide an example for how such work can yield valuable new insights regarding which approaches should be used under what conditions. We then conclude with several key recommendations, lessons learnt, and calls for change that would encourage and facilitate this line of work in the future, and accelerate our progress towards resolving the scientific debt that arises due to current PLM research practices (\S\ref{sec:explosive}). 

\vspace{-1mm}
\section{Analogy with Medicine and Parallels with PLM Research}\label{sec:medicine}
Consider the following analogy with clinical trials in medicine. A drug trial showing that drug A (taken 10 times a day) works better than drug B (taken only twice a day) would raise a few critical questions: Would drug A still work just as well if it is taken at a lower dose? Can we get the same results by increasing the dose of drug B? How well would each drug work under comparable conditions, and are there any particular trade-offs (\emph{e.g.,} at a lower dose, drug A works better than drug B; at a higher dose, drug B works better)? Answering these questions --- which requires disentangling the effects of the drug dose regimen --- would yield valuable insights, and facilitate more informed decisions over which drugs to produce at scale, and which drugs should be offered to which patients.

\vspace{-2mm}
\paragraph{Parallels with PLM research.} It is straightforward to see a parallel between this (flawed) drug trial setup with the way PLM research in NLP is done today. As each new PLM differs from earlier ones in multiple dimensions, it is increasingly harder to disentangle how much each component (\emph{e.g.,} objective function, model size, pre-training data amount) contributes to performance. This leaves us ill-equipped to answer questions like:
\vspace{-2mm}
\begin{itemizesquish}
    \item How well would earlier PLMs in the literature work if we augment them with the latest techniques, such as using the latest hyper-parameter settings or training them on more data? Would they match the performance of newer PLMs?
    \item To what extent should we attribute each PLM's performance to its key novelty (\emph{e.g.,} the \emph{bidirectional} masked language modelling loss for BERT), as opposed to other factors like the size of the model or the size of its training data?
    \item Which pre-training approaches should we use under considerations where efficiency considerations are paramount, such as for low-resource languages with limited amounts of monolingual data or under compute resource constraints?\footnote{Data efficiency considerations are also important for understanding and ``reverse-engineering'' human language learners, who are able to acquire language proficiency with much less amounts of data than current PLMs need \citep{hart_1995,dupoux_2018,cristia2019child,linzen-2020-accelerate}.}
\end{itemizesquish}

\vspace{-4mm}
\paragraph{Reasons for the scientific debt.} 

While answering the questions above would drive better-informed progress in the field, in practice there are multiple barriers to doing so; to some extent, these barriers account for why this scientific debt arises in the first place. These include (i) variations in the choice of hyper-parameters and experimental settings, which can result in a large variance in model performance \citep{bouthillier2019unreproducible,dodge2020fine}; (ii) the proprietary and opaque nature of many PLMs and their training datasets --- especially large-scale ones --- rendering standardization difficult; (iii) the increasing computational costs of large-scale PLMs, which increases the costs of running multiple pre-training experiments (\emph{e.g.,} with different model sizes or training data); and (iv) a strong emphasis in the field for achieving state-of-the-art results --- indirectly creating an incentive to spend all of one's compute, time, and effort to achieve the best results, albeit sometimes at the expense of scientific rigour, performing rigorous experiments under comparable conditions, tuning the baselines, and disentangling different possible sources of model improvements. We revisit these barriers, and outline our recommendations in \S\ref{sec:explosive}. 

\vspace{-2mm}
\paragraph{The costs and benefits of scientific debt, and \emph{why} we should address it.} In practice, the community goes into scientific debt because there are certain benefits of doing so. Indeed, by rapidly sharing and publishing new progress and state-of-the-art models --- even when we do \emph{not yet} fully understand how each factor contributes to the final performance of the model --- the community is able to share, use, and build on exciting findings (\emph{e.g.,} more accurate and faster models, etc.) much more quickly at a time of rapid progress. Yet on the other hand, accumulating too much scientific debt --- without a good plan to address it and eventually pay it off --- also carries an important risk: The lack of fair comparisons and proper ablations can lead the community down the wrong path, make sub-optimal choices, and waste precious community time, effort, and computational resources in the wrong direction. Paying off this scientific debt would crucially (i) enable the community to direct our collective efforts and resources into the research directions that matter the most for improving model performance, (ii) understand what factors enable PLMs' remarkable success today, and (iii) better comprehend the trade-offs between different approaches under various types of settings.


\vspace{-2mm}
\paragraph{Large variability in current PLMs.} To illustrate the extent of this issue, we summarize several key design choices behind some well-known PLMs in Table~\ref{tab:different} (Appendix~\ref{app:model_comparison}), revealing a large \emph{variability} in the key design choices (\emph{e.g.,} model size, training data corpus and size, subword pre-processing algorithm, pre-training task, etc.) behind each PLM. Some common patterns include scaling the model while also using different, often larger pre-training data, as well as using different training regimes altogether. As each design choice impacts model performance in different ways \citep{sennrich2019revisiting,jiao2019tinybert} --- combined with the fact that not all prior work conducted thorough ablations to understand how each component affects overall performance --- it has become increasingly difficult to understand \emph{why} a PLM outperforms the baselines, \emph{which} design choices should be used under what settings, and \emph{how much} of the improvement can be attributed to each work's novelty.

\section{Empirical Evidence: The Case of BERT}
\label{sec:empirical}
Having motivated the importance of better disentangling the impact of different design choices in PLM research, we conduct experiments in pursuit of this goal. These experiments serve to (i) further our understanding of the effectiveness of the masked language modelling \emph{objective} \citep{devlin2019bert}, compared to alternative approaches under comparable conditions; (ii) demonstrate how these experiments can yield new insights; and (iii) form the basis for our recommendations and lessons learnt for accelerating progress in this line of work.

At the time of its release, BERT \citep{devlin2019bert} attracted a lot of attention by virtue of its strong performance on many tasks, outperforming earlier PLMs like ELMo \citep{peters-etal-2018-deep} and GPT-1 \citep{radford2018improving} by substantial margins. At its core, BERT combines the following:
\begin{itemizesquish}
    \item Language model (LM) pre-training on large amounts of unlabelled data \citep[]{dai_le_2015,peters-etal-2018-deep,radford2018improving,howard-ruder-2018-universal}.
    \item Conducting \emph{whole-model fine-tuning} \citep{radford2018improving} on each downstream task, as opposed to only using the resulting contextual word representations (\emph{i.e.,} the neural model's frozen hidden state vectors) as features for a downstream task, as was done in the case of ELMo.
    \item Like the GPT-1 model, using a Transformers \citep{vaswani2017attention} architectural backbone, as opposed to (bidirectional) LSTMs \citep{hochreiter1997long} in the case of ELMo.
    \item Using a novel \emph{bidirectional} masked language modelling loss, which predicts the identity of each masked token by attending to \emph{both} its left and right context,\footnote{The bidirectional attention used within BERT is enabled by the use of Transformer architectures, which --- unlike LSTM models --- have no inherent directionality constraints.} rather than only the left/right context as in GPT-1 and ELMo pre-training.
\end{itemizesquish}
Based on the different components of BERT above, it combines previous known techniques with a key novelty: the masked language modelling objective, which --- unlike prior approaches like ELMo and GPT-1 --- enables it to leverage and fuse bidirectional context at pre-training. Nevertheless, BERT differs from its prior approaches on many \emph{other} factors (\emph{e.g.,} the pre-training data corpus and size, model size, using Transformers vs. LSTMs, length of the training cycle, tokenizer, etc.). This renders a principled comparison difficult, and makes it hard to \emph{isolate} the importance of the masked language modelling objective from other factors that also affect model performance. We therefore here ask:

\vspace{-3mm}
\begin{itemizesquish}
    \item To what extent can we attribute BERT's superior performance over its GPT-1 and ELMo baselines to its key novelty (\emph{i.e.,} the bidirectional masked LM loss), as opposed to other design choices?
    \item Can the baseline models achieve similar performance with BERT, if we augment them with a similar set of design choices that the BERT model used (\emph{e.g.,} whole model fine-tuning, using Transformers as opposed to LSTMs, etc.)?
    \item Can we come up with simpler variants of the baseline models, which can approximate the performance of more sophisticated approaches?
    \item How exactly would the findings change in the case where pre-training efficiency considerations are paramount (\emph{e.g.,} where there is a more limited amount of pre-training compute available)?
\end{itemizesquish}

\vspace{-5mm}
\subsection{Experimental Setup}\label{sec:experimental_setup}
We aim to isolate the importance of BERT's bidirectional masked LM objective, in comparison to two prior baselines: ELMo and GPT-1, under \emph{comparable} experimental conditions.
We compare our experimental setup with each model's original pre-training configuration in Table~\ref{tab:exp-param} (Appendix~\ref{app:hyper-params}).

\vspace{-2mm}
\paragraph{Training data.} For all three models, we use the original BERT pre-training data \citep{devlin2019bert}, containing a combination of Wikipedia\footnote{\label{note-license}All model and data license information is in Appendix~\ref{dataset-license}.} and BookCorpus~\cite{zhu2015aligning}. This dataset --- which is larger than either of the ELMo or GPT-1 pre-training dataset\footnote{As pre-training data size and quality have been shown to be an important factor of LM success \citep{liu2019roberta,hoffmann2022training}, we hypothesize that BERT's larger pre-training data is an important factor behind its success compared to prior approaches --- independently of the masked LM objective.} --- consists of $\sim 3.3$B words.

\vspace{-1mm}
\paragraph{Model.} 
We use a Transformer backbone for all three models, which has been shown to outperform LSTMs, and is more amenable to scaling to larger training datasets. Concretely, we use a BERT-Base architectural backbone, as implemented on HuggingFace\footref{note-license} \cite{wolf-etal-2020-transformers}, with $\sim 110$ million parameters. Whereas the underlying model architectures are identical across all models, the pre-training objective function is naturally tailored to each approach, \emph{e.g.,} masked LM for BERT, causal / left-to-right LM for GPT-1, and two independent causal LMs for ELMo: One operating in a left-to-right fashion, and another operating right-to-left. 

The implementation of the different pre-training objectives requires two changes. First, when applicable, we update the ``input mask'' function on each attention layer (\emph{e.g.,} using a standard causal attention mask in the case of GPT-1 to enforce a left-to-right directionality constraint). Second, we change the loss function to reflect each pre-training objective. For instance, we predict each next word conditional on its \emph{left} context for GPT-1; for BERT, we predict the $\sim 15\%$ masked words conditional on the (slightly corrupted) \emph{bidirectional} context. 
One key difference is that our BERT implementation excludes the next-sentence prediction (NSP) pre-training loss, in accordance with the findings of \citet{liu2019roberta}.\footnote{By using the same codebase for all three models, we eliminate confounds arising from minor technical differences, such as whether or not to use segment embeddings (as BERT does), what kind of positional encoding schemes are used, what vocabulary size and subword preprocessing algorithms are used, and how each batch of sequences is sampled.} All other variables (\emph{e.g.,} dataset, hyper-parameter choices, etc.) are kept identical across all models to facilitate a fair comparison.

\vspace{-1mm}
\paragraph{Pre-processing.} We use the same WordPiece tokenization as the original BERT model. We follow the procedure of \citet{liu2019roberta} for sampling sequences for each batch, where the input is constructed by repeatedly sampling multiple sentences until we reach the maximum sequence length of 512, while respecting document boundaries.

\vspace{-1mm}
\paragraph{Fine-tuning.} We use whole-model fine-tuning \citep{radford2018improving} for all models, which works better than the feature-based contextual word embedding approach of the original ELMo. As is standard practice, we take the top-layer contextual embedding of the [CLS] token to represent the whole sequence when fine-tuning BERT; for the left-to-right GPT-1 rerun, we take the top-layer contextual embedding of the \emph{last token}, where the model has observed the entire sequence.\footnote{By the same logic, we take the top-layer contextual embedding of the \emph{first} token to represent the whole sequence instead when fine-tuning the right-to-left GPT-1.} For each GLUE task, we run a grid search over 7 fine-tuning learning rates, 2 batch sizes, and 3 random seeds (Appendix~\ref{app:hyper-params}). We submit the best-performing model on the validation set to the GLUE leaderboard.

\vspace{-2mm}
\paragraph{ELMo rerun.} Following \citet{peters-etal-2018-deep}, we pre-train ELMo by independently pre-training two separate, causal / unidirectional models: a left-to-right one and a right-to-left one.\footnote{Note that there is no need for cross-model communication when pre-training the left-to-right and right-to-left ELMo models; hence this ELMo pre-training stage can be done completely in parallel, \emph{e.g.,} on two completely different machines.} At fine-tuning, \citet{peters-etal-2018-deep} combined the output layer of the left-to-right and right-to-left models, and used that combination as the representation of the whole sequence, based on which the fine-tuning cross-entropy loss is then calculated. In contrast, we employ a slight modification of ELMo where we first calculate the probability of each downstream task label under the left-to-right and right-to-left models, denoted as $p_{\boldsymbol{\theta}}^{\text{L2R}}(y \, | \, \mathbf{x})$, and $p_{\boldsymbol{\psi}}^{\text{R2L}}(y \, | \, \mathbf{x})$, respectively; $y$ denotes the downstream task label for a sequence $\mathbf{x}$, while $\boldsymbol{\theta}$ and $\boldsymbol{\psi}$ denote the parameters of the left-to-right and right-to-left models, respectively. The aim of ELMo fine-tuning is to find fine-tuned (FT) ELMo parameters $\left\{\boldsymbol{\theta}^{\star}_{\text{FT}},\boldsymbol{\psi}^{\star}_{\text{FT}}\right\}$:
\begin{align*}
& \boldsymbol{\theta}^{\star}_{\text{FT}},\boldsymbol{\psi}^{\star}_{\text{FT}} \stackrel{\text{def}}{=} \argmin_{\boldsymbol{\theta},\boldsymbol{\psi}} \sum_{(\mathbf{x}, \, y) \, \in \, D} \\ & - \log \left( \lambda \, p_{\boldsymbol{\theta}}^{\text{L2R}}(y \, | \, \mathbf{x}) + (1 - \lambda) \, p_{\boldsymbol{\psi}}^{\text{R2L}}(y \, | \, \mathbf{x}) \, \right).
\end{align*}
We simply set the interpolation coefficient $\lambda=0.5$ for all tasks. We find this variant of ELMo to perform better than the original ELMo formulation by a small margin ($\sim 0.5$\% in aggregate GLUE validation performance) in our preliminary experiments.

\vspace{-2mm}
\paragraph{Hyper-parameters.} We summarize the pre-training and fine-tuning hyper-parameters for each model in Appendix~\ref{app:hyper-params}. We spend a similar amount of compute in tuning the hyper-parameters of each model, facilitating a fair comparison across models.

\vspace{-2mm}
\paragraph{Evaluation Tasks.} We evaluate each model on the GLUE classification tasks \citep{wang-etal-2018-glue}, as done in the original BERT paper. We leave other evaluation benchmarks like SuperGLUE \citep{wang2019superglue} to future work, although due to BERT's bidirectional nature, conducting generative evaluation that requires sampling text from BERT is non-trivial \citep{wang-cho-2019-bert,goyal-etal-2022}.

\vspace{-2mm}
\paragraph{Compute.} Pre-training each model took 5 days with 8 V100 GPUs, while it took roughly a day to run GLUE fine-tuning with 8 GPUs (42 fine-tuning hyper-parameters for each \{model, task\}, \S\ref{sec:experimental_setup}). All in all, we used 8700 GPU hours for pre-training and 1100 GPU hours for fine-tuning all models.

\subsection{Empirical Findings}
We summarize the GLUE test set results in Table~\ref{tab:test-1M}, based on which we remark on four observations.
\vspace{-2mm}
\begin{itemizesquish}
    \item Under comparable conditions, the test GLUE performance of our variant of ELMo is at $76.8$\%, representing a relatively small 1.2\% gap with the standard BERT rerun ($78$\%). This improved performance represents a vast, $>6$\% improvement from the original ELMo's reported result of $70.3$\%; we attribute this gap to the use of a larger, BERT-equivalent pre-training data, a Transformer backbone, and whole model fine-tuning. We also see a small gain of our GPT-1 Rerun --- an improvement we attribute to a larger and better pre-training dataset than the original model. Hence, we conclude that most of the substantial $>8$\% gap between the original BERT and ELMo results can, in fact, be attributed to using Transformers rather than LSTMs, conducting whole model fine-tuning, and using a larger pre-training data, rather than BERT's bidirectional masked LM loss in and of itself. Altogether, this result reaffirms how augmenting the baselines with a similar set of techniques and advances as later generation models can substantially improve their performance, and yield results that are close to those of more recent models \citep[\emph{inter alia}]{melis_etal_2018,merity_2019,lei-2021-attention}. 
    \item However, there remains a larger gap between the causal / unidirectional PLMs and BERT; this holds for both the left-to-right ($2.6$\% worse than BERT) and the right-to-left ($3.4$\% worse) models. Note, however, that this $2.6$\% gap under comparable conditions is smaller than the original reported GPT-1 result, which had a $>4\%$ gap with the original BERT --- a result we attribute to the smaller dataset used to pre-train the original GPT-1. These findings further emphasize the importance of comparing the baselines and more recent PLMs under comparable conditions.
    \item Although the left-to-right and right-to-left models still lag behind BERT, a simple \emph{ensemble} of two independently pre-trained and fine-tuned left-to-right and right-to-left models can nevertheless approach BERT's and ELMo's performance (76.3\% for the ensemble, 78\% and 76.8\% for BERT and ELMo reruns, respectively). Remarkably, we do not observe the same gains when ensembling two left-to-right models from different random seeds, suggesting that ensembling unidirectional models with different directionalities is crucial for performance. All in all, these results highlight the need to explore \emph{simpler baselines}, which can approximate the performance of more sophisticated approaches.\footnote{Ensembling independently pre-trained and fine-tuned left-to-right and right-to-left models is a \emph{late-fusion} approach, without the need for ELMo's \emph{joint} fine-tuning stage.} 
    \item As shown in Table~\ref{tab:test-200k}, when efficiency considerations are paramount (\emph{i.e.,} where each model is only pre-trained for 200k steps, and not the full 1M), the performance gap between the left-to-right GPT-1 rerun and BERT nearly vanishes (0.5\% gap, as opposed to a 2.6\% gap in the 1M-pre-training-steps setup). We attribute this to the fact that BERT only uses $\sim15$\% of the tokens in a batch as the masked LM target, whereas the left-to-right LM can leverage all 100\% of the tokens as pre-training supervision in a similar fashion as Electra \citep{clark2020electra}, hence resulting in more efficient learning. Remarkably, despite its simplicity, an ensemble of independently-pre-trained-and-fine-tuned left-to-right and right-to-left models \textbf{outperforms} the BERT model by 1.1\% in this efficient learning scenario. This finding (i) suggests that approaches that work best in the high-data/compute scenario may not necessarily transfer to cases where efficiency considerations are paramount; and (ii) highlights the need to train, evaluate, and ultimately \emph{benchmark} models in efficient learning scenarios, such as in languages where monolingual data are not abundant or in cases where there is only a limited amount of compute available for pre-training.
\end{itemizesquish}

\vspace{-3mm}
\paragraph{Validation set results with error bars.} To preserve test set integrity, we only submitted the single-best validation model to the test set (Table~\ref{tab:test-1M}). In Appendix~\ref{app:results_with_stdev}, we report the validation set performance that includes error bars over three different random seeds, which broadly show the same trend.

\begin{table*}[ht!]
    \centering
    \small
    \resizebox{0.99\textwidth}{!}{
    \begin{tabular}{@{}llllllllll@{}}
    \toprule
        Model & CoLA & MNLI(-m) & MRPC & QNLI & QQP & RTE & SST-2 & STS-B & Avg \\
    \midrule
    \multicolumn{10}{c}{\textbf{Original published results}} \\ 
    \midrule
        BERT & 52.1 & 84.6 & 88.9 & 90.5 & 71.2 & 66.4 &  93.5 &  85.8 & 79.1 \\
        BERT Large & 60.5 & 86.7 & 89.3 & 92.7 & 72.1 & 70.1 & 94.9 & 86.5 & 81.6 \\
    \midrule
        GPT-1 & 45.4 & 82.1 & 82.3 & 87.4 & 70.3 & 56.0 & 91.3 & 80.0 & 74.4 \\
    \midrule
        BiLSTM + ELMo + Attn & 36.0 & 76.4 & 84.9& 79.8 & 64.8 & 56.8 &  90.4 & 73.3 & 70.3 \\
    \midrule 
    \multicolumn{10}{c}{\textbf{Our replication with proper controls \& comparable experimental conditions}} \\ 
    \midrule 
        BERT Rerun & 50.8 & 84.5 & 89.0 & 90.5 & 71.0 & 61.0 & 93.1 & 84.4 & 78.0 \\ 
    \midrule
        Comparable GPT-1 Rerun - L2R & 41.6 & 87.4 & 84.7 & 86.6 & 68.8 & 62.9 & 91.8 & 79.3 & 75.4 \\ 
        Comparable GPT-1 Rerun - R2L & 42.5 & 82.0 & 85.5 & 88.3 & 69.1 & 57.6 & 92.8 & 79.1 & 74.6 \\ 
    \midrule
        Comparable ELMo-variant Rerun  & 46.8 & 83.6 & 85.8 & 89.9 & 70.8 & 61.9 & 93.1 & 82.1 & 76.8 \\ 
        Ensemble of Comparable GPT-1: L2R + R2L & 45.1 & 83.7 & 85.8 & 88.9 & 70.8 & 62.4 & 92.9 & 81.0 & 76.3 \\ 
        Ensemble of Comparable GPT-1: L2R + L2R & 42.4 & 83.5 & 85.1 & 87.8 & 70.0 & 63.1 & 93.1 & 79.9 & 75.6 \\ 
       
    \bottomrule
    \end{tabular}}
    \vspace{-1mm}
    \caption{GLUE \textbf{test} results. We use F1 scores for MRPC and QQP, Matthew's Correlation for CoLA, SpearmanR for STS-B, and accuracy for the rest; all models are pre-trained with the same batch size \& compute (1M steps).}
    \label{tab:test-1M}
\end{table*}

\begin{table*}[ht!] 
    \centering
    \small
    \resizebox{0.99\textwidth}{!}{
    \begin{tabular}{@{}lllllllllll@{}}
    \toprule
        Model & CoLA & MNLI(-m) & MRPC & QNLI & QQP & RTE & SST-2 & STS-B & Avg \\
    \midrule
        BERT Rerun & 43.8 & 80.9 & 86.4 & 87.9 & 69.3 & 59.3 & 90.0 & 80.4 & 74.8 \\ 
    \midrule
        Comparable GPT-1 Rerun: L2R  & 43.5 & 80.3 & 84.1 & 86.3 & 68.4 & 63.0 & 91.0 & 77.8 & 74.3
        \\ 
        Comparable GPT-1 Rerun: R2L  & 36.2 & 80.6 & 82.4 & 88.2 & 68.7 & 53.7 & 93.0 & 77.8 & 72.6 \\ 
    \midrule
        Ensemble of GPT-1 Rerun: L2R + R2L & 45.1 & 82.6 & 84.4 & 88.3 & 70.8 & 62.9 & 93.5 & 79.9 & 75.9 \\ 
    \bottomrule
    \end{tabular}}
    \vspace{-1mm}
    \caption{GLUE test set results using the pre-trained model, after training for 200,000 steps followed by fine-tuning.}
    \label{tab:test-200k}
\vspace{-2mm}  
\end{table*}

\section{Paying Off the Scientific Debt: Recommendations and Lessons Learnt}
\label{sec:explosive}
We proceed to outline several key recommendations and lessons learnt for encouraging, incentivizing, and accelerating progress in this line of work.

\vspace{-1mm}
\paragraph{Establish standard, publicly available pre-training corpora at multiple data scales.} As seen in \S\ref{sec:empirical}, the size and quality of the pre-training data is an important driver behind model performance \citep{liu2019roberta,hoffmann2022training}, which makes a rigorous comparison between different PLMs difficult. Hence, our first recommendation is to establish standard pre-training corpora that are publicly available.\footnote{The establishment of standardized corpora (along with their corresponding training / validation / test splits) have been a standard feature of statistical NLP for the last decades. To that end, researchers would compare models that are trained and evaluated on the same datasets to enable a fair comparison, such as the Penn Treebank \citep{marcus-etal-1993-building} for parsing or SQuAD for question answering \citep{rajpurkar-etal-2016-squad}. But pre-trained foundation models \citep{foundation_models} introduce an additional confound: two PLMs that are fine-tuned on the exact same task and dataset can have different performance simply because one of them is \emph{pre-trained} on more data, irrespective of the novelty of each PLM.} We further recommend releasing the pre-training corpora under multiple data scales, as approaches that work best under strict compute or data resource requirements may be different from the case where there is a large amount of compute and data available \citep[\S\ref{sec:empirical},][]{clark2020electra,treviso_etal_2022}. Note that this does \emph{not} mean that we are discouraging the use of non-standard or even-larger corpora than those that are publicly available. On the contrary, researchers \emph{should} continue to push the boundaries of what is possible by training on more, better quality, and more recent data. In such cases, we recommend researchers to \emph{also} release versions of their models that are trained on the standard pre-training corpora --- above and beyond the version trained on proprietary \& large-scale data that would presumably be necessary to achieve a new state-of-the-art --- in order to facilitate a fair and principled comparison with prior work. We encourage the community to \emph{continually} release new standardized pre-training datasets as time passes to avoid the effect of pre-training data staleness \citep{mind_the_gap}.

\vspace{-2mm}
\paragraph{Explicitly delineate the different types of contributions behind each work, including both the key novelty and engineering contributions.} We recommend that PLM research explicitly state the key novelty behind each work (\emph{e.g.,} the bidirectional masked LM loss for BERT), delineate and explicitly state other contributions (including engineering ones) and design choices that can impact performance, and outline how these differ from prior work (\emph{e.g.,} better model partitioning for training larger models on multiple devices, better filtering of the training data, more extensive hyper-parameter tuning, etc.). Combined with strong baselines and extensive ablations (see below), this will enable us to better understand \emph{how much} of the performance gains can be attributed to each factor, including the key novelty behind the approach.

\vspace{-2mm}
\paragraph{Invest comparable effort into tuning both the baselines \emph{and} the newly-proposed models.} In practice, many of the contributions (\emph{e.g.,} better hyper-parameters, better data, etc.) would also be applicable to the baselines. We recommend each PLM work to discern which of their design choices can \emph{also} be applied to the baselines, and apply those techniques in order to create stronger baselines that may nevertheless rival the performance of more recent models \citep[\S\ref{sec:empirical},][]{melis_etal_2018,lei-2021-attention}.

\vspace{-2mm}
\paragraph{More extensive ablation studies.} When proposing multiple contributions at once (as many PLM papers do), we recommend conducting as many ablation studies as is feasible to isolate the impact of each component under comparable conditions. In light of recent trends where models --- including open-sourced ones --- are publicly released without technical reports or papers that outline technical details regarding model evaluation and benchmarking~\cite{alpaca, vicuna2023}, we argue that our recommendation for conducting more thorough evaluations is even more critical.

\vspace{-2mm}
\paragraph{Better credit assignment is needed.} As shown in Table~\ref{tab:test-1M}, the vast gap between BERT and ELMo can nearly be bridged by using (i) the same (larger) pre-training data, (ii) Transformer architectures, and (iii) whole model fine-tuning; all of which were already used and proposed by the GPT-1 model. As these techniques account for a more significant chunk of the performance difference than the bidirectional masked LM loss, disentangling each factor's contribution thus provides an opportunity to conduct better credit assignment in the field.

\vspace{-2mm}
\paragraph{Strike a balance between pushing the state-of-the-art and advancing our scientific understanding.} 
In some sense, recent rapid progress is made possible by a strong emphasis on building the next state-of-the-art PLMs and foundation models, although it comes at a cost of understanding --- from the scientific point of view --- where the performance improvements are coming from, and which techniques work best under what circumstances. 
We argue that \emph{both} lines of work --- one that pushes the state-of-the-art at breakneck speed and through all available means, and another that aims to resolve the scientific and technical debt by disentangling the impact of multiple factors of model improvements (which we argue is still currently underrepresented in the field) --- should be conducted, encouraged, and rewarded within the field. 
We outline two concrete recommendations for striking a better balance between the two lines of work. 
First, public release of PLMs or their downstream applications should be \emph{promptly} accompanied by a technical description of the model, ideally in the form of a technical report or a scientific paper.\footnote{In some cases, the relevant technical descriptions were not promptly released following model release, such as the community project BLOOM \cite{scao2022bloom} or ChatGPT (whose blog post does not cover much technical detail.).} 
This would enable the community to better understand the key component behind these models' success, allow future work to replicate the results, and promptly disentangle the different components behind model improvements. 
Second, we as a community should not necessarily expect \emph{both} types of contributions under the same paper. 
Just like how the cleaning up of technical debt happens \emph{after} the initial code has been written, it is often the case that prior work that resolves the scientific debt through principled comparisons was only conducted after substantial progress in advancing the state-of-the-art (often through all available means for improving model performance) had been made. 
We should, however, encourage the community to conduct such understanding line of work promptly after major milestones or exciting results.

\vspace{-1mm}
\paragraph{Reward and encourage a line of work that focuses on understanding (not just those that chase a new state-of-the-art), even when they are imperfect.} 
The current, rapid pace of the field provides an incentive to spend one's (finite) computational resources and effort for building the next state-of-the-art, albeit at the expense of scientific rigour and principled comparisons. Given a finite amount of compute, there is arguably more incentive in tuning one's proposed approach through all possible means (\emph{e.g.,} using larger datasets and larger models, training for longer, etc.), topping the leaderboards, and publishing the paper, even if this leaves no computational resources to tune the baselines and conduct rigorous ablations. Furthermore, the rapidly increasing cost of training ever-larger PLMs means that any principled comparisons are most likely imperfect (\S\ref{sec:limitations}) --- \emph{e.g.,} how do our findings in \S\ref{sec:empirical} change with models that are trained for longer, like RoBERTa? Or with encoder-decoder models like T5? Or in other languages? 
 Indeed, our experiments in \S\ref{sec:empirical} are fairly narrow in scope, involving only three non-recent models (BERT, ELMo, GPT-1) and a training dataset that is small by today's standards. Yet due to the rigorous hyper-parameter tuning of all three models, conducting these principled comparisons required an enormous amount of compute resources --- equivalent to training 10 BERTs from scratch. This cost would have been even higher with the inclusion more models, languages, and larger datasets. On this point, we remark that doing such principled comparisons --- even when they are limited in scope and done on smaller models --- \emph{still} contributes towards paying off the scientific debt, better understanding where our current progress is coming from, and deriving valuable insights that can contribute to the development of next generation PLMs. We additionally call on those in our community who serve as reviewers to recognize and reward these types of research contributions, which are \emph{complementary} (if perhaps equally important) to a parallel line of work that pushes the state-of-the-art in PLM research through all possible means.

\vspace{-2mm}
 \paragraph{We need more comprehensive PLM scaling laws.} Our experiments and recommendations still leave a major open question: How can we scale these kinds of investigations to much larger PLMs, which are much more computationally expensive? To that end, \textbf{scaling laws} \citep{kaplan2020scaling,hoffmann2022training} provide an account of how PLM performance changes with respect to different factors, allowing us to accurately extrapolate that a PLM with X parameters and Y training steps should achieve a perplexity of Z. However, we argue that current scaling laws are still overly narrow in scope: Concretely, existing scaling laws often only apply to decoder-only / unidirectional PLMs, and only provide an account of how their performance changes with respect to (i) model size and (ii) the number of training tokens. We call on the community to develop more comprehensive scaling laws that take into account and characterize how other factors impact LM performance and downstream behavior, including how model performance and behavior change with respect to the choice of the objective function and model hyper-parameters, and the quality of the pre-training data. The existence of such scaling laws --- which can happen by pooling community data on various PLM pre-training runs and their corresponding perplexity and downstream performance --- would allow other researchers to accurately \emph{extrapolate} how their findings would generalize to other PLM model sizes, objective functions, etc. Most importantly, comprehensive scaling laws can disentangle and \emph{quantify} how these different factors contribute to determine the final model performance under various experimental conditions.

 \paragraph{How conducting rigorous experiments and ablation studies can lead to new state-of-the-art results.} Lastly, we argue that conducting rigorous experiments and ablation studies for paying off the scientific debt should \emph{not} necessarily come at the expense of achieving a new state-of-the-art. In contrast, doing so can be a key ingredient for building the next state-of-the-art PLMs. In 2020, \citet{kaplan2020scaling} proposed a seminal scaling law that showed how larger PLMs are more sample-efficient, and that one should always increase model size when increasing the pre-training compute budget, leading the community to develop ever-larger PLMs in response \citep[\emph{inter alia}]{rae2021scaling,smith_etal_2022}. Nevertheless, subsequent rigorous experiments from \citet{hoffmann2022training} demonstrated that the optimal pre-training compute allocation should, in fact, \emph{also} be scaled in another dimension: The amount of pre-training data that the model is trained on. This insight was then used to build smaller, more efficient, and cheaper-to-run PLMs that, at the time of its release, achieved new state-of-the-art results that outperformed much larger PLMs that were under-trained in comparison. Going forward, we conjecture that rigorous experiments and ablation studies that look at factors \emph{above and beyond} model size and data quantity, such as the \emph{quality} of the pre-training data, the exact hyper-parameters, the pre-training objective, etc., will not only be useful to understand how these factors improve performance and thus pay off the scientific debt, but also form a key ingredient for building the next generation of better PLMs.

\vspace{-2mm}
\section{Related Work}
A number of prior work has made progress in disentangling the impact of different language modelling pre-training objectives by conducting principled ablation studies under comparable experimental conditions \citep[\emph{inter alia}]{unilm,raffel2020exploring,tay2022unifying,artetxe_2022}. However, some of the recently released models do not provide any technical details on how they are trained, such as ChatGPT\footnote{The model inference is accessible via OpenAI API.} and GPT-4~\cite{bubeck2023sparks}. We discuss these in an extended related work section (Appendix~\ref{app:related_work}), but briefly remark on how our findings complement theirs. First, we revisit and augment ELMo --- which incorporates a degree of bidirectionality at fine-tuning (albeit not at pre-training) --- with Transformers and whole model fine-tuning, facilitating a fair comparison with BERT. We show that the resulting ELMo achieves competitive performance with BERT on GLUE; to our knowledge, no such ELMo baseline --- or an even simpler ensemble of a left-to-right and right-to-left PLM --- was explored in prior work. 

While our work shares several similarities with~\cite{melis_etal_2018}, our work differs by virtue of being a position paper that focuses on an important issue in the field (\emph{i.e.,} the lack of fair comparisons between past PLMs), and chart the way forward for mitigating this issue. Our experiments in this work mostly aim to provide an example of this issue in action, and form the basis for some of the lessons learnt and recommendations that we outline in \S\ref{sec:explosive}. Unlike \citet{melis_etal_2018}, our experiments are not designed to achieve new state-of-the-art results. Moreover, above and beyond our empirical contributions, we outline key recommendations that would encourage and incentivize this line of work; we hope that these recommendations would be adopted by the broader community, with the aim of accelerating progress towards resolving the scientific debt in foundation model research.

\vspace{-2mm}
\section{Conclusion}
Recent rapid progress within the PLM literature has led to tremendous advances within NLP. Despite this progress, current PLM research practices that change multiple different things at once --- often without proper ablation studies and conducting principled comparisons that disentangle the impact of different components --- have introduced certain issues that we call ``scientific debt''. Through experiments that disentangle the contribution of BERT's bidirectional masked LM objective through principled comparison with prior work, we demonstrate how asking ``\emph{which factors contribute the most to the model performance that we observe today}?'' can lead to valuable new insights, including the existence of simple yet stronger-than-expected and more efficient baselines. We outlined several recommendations that would encourage and incentivize this line of work that aims to better understand how each factor contributes to the rapid progress of our PLMs today, and better address the ongoing issue of accumulating scientific debt within our current PLM research literature.

\section{Limitations}\label{sec:limitations}
Our work has the following limitations.

\vspace{-1mm}
\paragraph{Comparisons with more recent models.} In \S\ref{sec:empirical}, we conducted a principled comparison between BERT, ELMo, and GPT-1 under comparable experimental conditions. This comparison notably excludes more recent models that benefit from more parameters, larger training data, or different loss functions, such as RoBERTa, Electra, and T5. Due to the even-higher cost of pre-training these more recent models, we leave a principled comparison that includes these models to future work, although we identified the development of more comprehensive PLM scaling laws as a promising future research direction that would allow us to extrapolate how our findings would generalize to different pre-training data sizes, objective functions, etc. (\S\ref{sec:explosive}).

\vspace{-1mm}
\paragraph{Interaction between different factors.} In \S\ref{sec:empirical}, we have conducted a principled comparison by varying only the pre-training objective function and the length of model training, whilst keeping all the other variables constant. In practice, however, the exact choice of these different control variables (\emph{e.g.,} what positional encodings to use, how we pre-process the data, etc.) can \emph{interact} and affect the findings in a material way. It is conceivable --- and rather likely --- that our findings on the performance gap between BERT, ELMo, and GPT-1 may change under different experimental settings.

\vspace{-1mm}
\paragraph{Simulated efficient learning scenario.} Our efficient learning scenario in \S\ref{sec:empirical} constitutes a simulated one, where we artificially limit the number of updates to 200,000 steps (as opposed to 1M steps in the full setting). We leave the extension to more realistic efficient learning scenarios, such as in languages where there is only a limited number of monolingual data, or where there is a hard limit on what pre-training computational resources we can use (\emph{e.g.,} 1 GPU for 3 days), to future work.

\vspace{-1mm}
\paragraph{Extension to multi-lingual settings.} Our experiments are thus far conducted only in English. We leave the extension to other languages --- including low-resource languages with only a limited amount of monolingual data as a realistic and necessary benchmark of efficient learning --- to future work.

\paragraph{The increasing prevalence of closed-source / proprietary PLMs.} Despite our recommendations and calls for change, we acknowledge the fact that recent PLM trends have shifted more towards proprietary and closed-source models --- a development we attribute to the rapidly increasing commercialization potential of this technology. Under this trend, very little is known about how each PLM is developed, as the vast majority of the technical details (\emph{e.g.,} the amount and source of the pre-training data, the data filtering strategy, the size and hyper-parameters of the model, how the model is implemented, etc.) are kept proprietary. While these trends may mean that our recommendations are more unlikely to be adopted by proprietary PLMs, we argue that our position paper and recommendations are still important (if not even more so) for two reasons. First, open-sourced community models, such as BLOOM \citep{scao2022bloom}, OPT \citep{zhang2022opt}, and Alpaca \citep{alpaca}, are gaining traction, and have rapidly narrowed the gap with proprietary models. This progress reflects the community's strong desire to have open-sourced models that can rival proprietary ones in terms of model quality. The rise of these open-sourced models thus gives rise to the question: How can these community-driven models help the community pay off our scientific debt? To that end, our recommendations provide concrete and actionable steps in this direction. For instance, our recommendations call for standardizing the pre-training dataset, which has not yet been done thus far, even though there are plausible, open-sourced datasets that can be used for doing so. Furthermore, we also encourage the community to release the full evaluation results of their models, alongside the relevant hyper-parameter information, etc., such that we can \emph{collectively} build a more comprehensive scaling law through crowd-sourcing (\S\ref{sec:explosive}). Second, prior work that conducts extensive ablation studies and rigorous experiments \citep[\emph{inter alia}]{raffel2020exploring,sun-iyyer-2021-revisiting} remains the exception, rather than the rule. Our position paper includes a call for change that will make it \emph{easier} to pay off this scientific debt going forward, which is ever-more important in light of impressive progress from both proprietary and open-sourced PLMs.

\vspace{-1mm}
\section*{Ethical Considerations}
Our experiments replicate prior work under comparable experimental conditions. For this reason, we do not expect our work to introduce any novel ethical issues, although our experiments may inherit a similar set of issues concerning PLM (especially large-scale ones), as outlined by various prior work \citep[\emph{inter alia}]{gehman-etal-2020-realtoxicityprompts,stochastic_parrots,rae2021scaling,dinan_etal_2021,foundation_models,kenton_etal_2021,weidinger_etal_2021}. We remark, however, that conducting these principled comparisons across different models --- which requires a degree of hyper-parameter tuning for each model (both at pre-training and fine-tuning stages) in order to enable a fair comparison --- requires a large number of computational resources, which may contribute to increased carbon emissions \citep{strubell-etal-2019,patterson-etal-2021}.

\section*{Acknowledgement}
We thank Chris Dyer, John T. Hale, and Laura Rimell at DeepMind, Noah A. Smith at the University of Washington \& Allen Institute for Artificial Intelligence, and Qi Huang at Bloomberg for their valuable insights, feedback, and suggestions.

\bibliography{custom}
\bibliographystyle{acl_natbib}

\appendix

\newpage

\section{Model Comparison}\label{app:model_comparison}
In Table~\ref{tab:different}, we outline a summary of various aspects of some commonly-used PLMs that have been proposed to date, summarizing the size of the model, the training data and its size, the text pre-processing scheme, the pre-training objective, and some hyper-parameter details. This table reveals a large variation in the design choices of these PLMs, hence rendering it difficult to conduct apple-to-apple comparisons between different approaches. Common patterns include scaling the model while also using different (often larger) pre-training data, as well as using different training regimes altogether. Each design choice impacts model performance in different ways \citep{sennrich2019revisiting,jiao2019tinybert}, emphasizing the importance of conducting thorough ablations and apple-to-apple comparisons.

\begin{table*}[!ht]\centering
\resizebox{0.99\textwidth}{!}{
\begin{tabular}{cccccc}
\toprule
\textbf{Model Size} & \textbf{Training Data (Size) }& \textbf{Text Pre-processing} & \textbf{Pre-training Objective} & \textbf{Setup} \\
\midrule
\multicolumn{5}{c}{BERT \cite{devlin2019bert}
$\diamondsuit$} \\
\midrule
\makecell{Smallest: 110M\\Largest: 340M} 
    &\makecell{English Wikipedia, Book Corpus\\(16 GB)} 
    &\makecell{Wordpiece\\(30k tokens)} 
    &\makecell{ MLM and NSP \\ simultaneously } 
    &\makecell{\textbf{Optimizers}: 1e-4, Adam, Linear Decay\\
        \textbf{Batch Size, Max Sequence Length}: 256, 512\\
        \textbf{Steps/Epochs}: 1,000,000}\\
\midrule
\multicolumn{5}{c}{RoBERTa \cite{liu2019roberta}
$\diamondsuit$} \\
\midrule
\makecell{Smallest: 125M\\Largest: 355M} 
    &\makecell{English Wikipedia, Book Corpus\\CCNews, OpenWebText, Stories} 
    &\makecell{BPE\\(50k tokens)} 
    &\makecell{ MLM } &\makecell{\textbf{Learning Rate}: (7e-4, 1e-4, 1e-3), Adam, Linear decay \\ 
        \textbf{Batch Size, Max Sequence Length}: 256, 512 \\ 
        \textbf{Steps/Epochs}: 1,000,000} \\
\midrule
\multicolumn{5}{c}{Megatron-BERT \cite{shoeybi2019megatron}
$\diamondsuit$} \\
\midrule
\makecell{Smallest: 336M\\Largest: 3.9B} 
    &\makecell{
        Wikipedia, RealStory \\
        Book Corpus, CC-Stories \\ 
        OpenWebText \\
        (174 GB)} 
    &\makecell{Wordpiece\\(30k tokens)} 
    &\makecell{ MLM and NSP \\ simultaneously } 
    &\makecell{\textbf{Optimizers}: 1e-4, Adam, Linear Decay \\
        \textbf{Batch Size}: 1024 \\
        \textbf{Steps/Epochs}: 2,000,000} \\ 
\midrule
\multicolumn{5}{c}{GPT-1 \cite{radford2018improving} $\spadesuit$} \\
\midrule
\makecell{117M} 
    &\makecell{1B Word Benchmark\\Book Corpus\\($\sim$ 9GB)} 
    &\makecell{BPE\\(40k tokens)} 
    &\makecell{CLM} 
    &\makecell{\textbf{Optimizer}:2.5e-4, 
        \\Adam, cosine annealing scheduler\\
        \textbf{Batch Size, Context Size}:  64, 512\\
        \textbf{Steps/Epochs}: 100 epochs}\\
\midrule
\multicolumn{5}{c}{GPT-2 \cite{radford2019language} $\spadesuit$} \\
\midrule
\makecell{Smallest: 110M\\Largest: 1542M} 
    &\makecell{WebText\\(40 GB)} &\makecell{BPE\\ (52k tokens)} 
    &\makecell{ CLM } 
    &\makecell{\textbf{Optimizer}: tuned, unknown hyperparameter \\
        \textbf{Batch Size, Context Size}: 512, 1024}\\
\midrule
\multicolumn{5}{c}{GPT-3 \cite{brown2020language} $\spadesuit$} \\
\midrule
\makecell{Smallest: 125M\\Largest: 172B} 
    &\makecell{Expanded WebText\\Filtered CommonCrawl\\Internet Book Corpora\\English Wikipedia \\(499B tokens)} 
    &\makecell{BPE \\(52k tokens)} 
    &\makecell{ CLM } 
    &\makecell{
        \textbf{Optimizer}: Adam, Linear Warmup\\
        \textbf{Batch Size, Context Size}: Dynamic, 2048} \\
\midrule
\multicolumn{5}{c}{Megatron-GPT \cite{shoeybi2019megatron} $\spadesuit$} \\
\midrule
\makecell{Smallest: 355M\\Largest: 8.3B} 
    &\makecell{ 
        Wikipedia, RealStory, \\
        Book Corpus, CC-Stories \\
        OpenWebText \\ 
        (174 GB) } 
    &\makecell{Follow GPT-2} 
    &\makecell{ CLM } 
    &\makecell{
        \textbf{Optimizer}: 1.5e-4, Adam, \\ warmup + cosine decay \\
        \textbf{Batch Size, Context size}: 512, 1024 \\
        \textbf{Steps/Epochs}: 300k steps} \\
\midrule
\multicolumn{5}{c}{OPT \cite{zhang2022opt} $\spadesuit$} \\
\midrule
\makecell{Smallest: 125M\\Largest: 175B} 
    &\makecell{ 
        English Wikipedia, CC-Stories \\
        The Pile (deduped) \\
        PushShift.io Reddit \\
        CCNewsV2 \\
        (180B Tokens, 800GB) } 
    &\makecell{Follow GPT-2} 
    &\makecell{ CLM } 
    &\makecell{
        \textbf{Optimizer}: Adam, Linear scheduling \\
        \textbf{Batch Size, Context Size}: 0.5M to 4M tokens, 2048} \\
\midrule
\multicolumn{5}{c}{T5 \cite{raffel2020exploring} $\diamondsuit$$\spadesuit$} \\
\midrule
\makecell{Smallest: 220M\\Largest: 11B} 
    &\makecell{
        C4 filtered \\
        (750 GB)} 
    &\makecell{Wordpiece\\(32k tokens)}
    &\makecell{
        MLM on encoder/decoder, \\
        continued with prompted tasks} 
    &\makecell{
        \textbf{Optimizer}: 0.01, Adafactor, insqrt scheduler\\ 
        \textbf{Batch Size, Max Sequence Length}: 128, 512\\
        \textbf{Steps/Epochs}: 524,288} \\
\midrule
\multicolumn{5}{c}{BART \cite{lewis-etal-2020-bart} $\diamondsuit$$\spadesuit$} \\
\midrule
\makecell{Smallest: 140M\\Largest: 400M} 
    &\makecell{
        English Wikipedia, Book Corpus \\
CCNews, OpenWebText, Stories \\
        (160 GB)} 
    &\makecell{BPE\\(50k tokens)} 
    &\makecell{
        CLM with corrupted \\
        encoder input } 
    &\makecell{
        \textbf{Optimizer}: tuned \\ 
        \textbf{Batch Size, Context Size}: 8000, tuned\\
        \textbf{Steps/Epochs}: 500,000} \\
\midrule
\multicolumn{5}{c}{Gopher \cite{lewis-etal-2020-bart} $\diamondsuit$$\spadesuit$} \\
\midrule
\makecell{Smallest: 44M\\Largest: 280B } 
    &\makecell{
        Massive Web \\ 
        Books, C4, News, \\ 
        GitHub, Wikipedia \\ 
        (11 TB)} 
    &\makecell{BPE\\(32k tokens)} 
    &\makecell{ CLM } 
    &\makecell{User RMSNorm and relative positional encoding \\ 
        \textbf{Optimizer} Adam (different LR) \\ 
        \textbf{Batch Size, Context Size}: 0.25 M to 6M \\
            (depends on the model size), 2048\\
        \textbf{Steps/Epochs}: unknown} \\
\midrule
\multicolumn{5}{c}{Chinchilla \cite{lewis-etal-2020-bart} $\diamondsuit$$\spadesuit$} \\
\midrule
\makecell{70B} 
    &\makecell{Same as Gopher} 
    &\makecell{BPE w/o NFKC-norm\\(32k tokens)} 
    &\makecell{ CLM } 
    &\makecell{Similar to Gopher, \\
        but use AdamW as its optimizer} \\
\midrule
\multicolumn{5}{c}{BLOOM \cite{scao2022bloom} $\diamondsuit$} \\
\midrule
\makecell{Smallest: 560M\\Largest: 176B} &\makecell{ROOTS Corpus\\ (1.6 TB)} 
    &\makecell{Byte-level BPE\\(250k tokens)} 
    &\makecell{ CLM } 
    &\makecell{
        Alibi positional embedding\\
        float16 precision training\\
        \textbf{Batch Size, Max Sequence Length}: 2048, 2048\\} \\
\bottomrule
\end{tabular}}
\caption{
    Comparison pre-training setup between popular language models. 
    $\diamondsuit$$\spadesuit$: Encoder-Decoder Transformer. 
    $\diamondsuit$: Encoder Transformer 
    $\spadesuit$: Decoder Transformer. 
    MLM: Masked Language Modelling, NSP: Next Sentence Prediction, CLM: Causal Language Modelling. Note that what we put here is based on the paper for each language model.}
\label{tab:different}
\end{table*}

\section{Detailed Hyper-Parameters}\label{app:hyper-params}
When conducting experiments, we follow BERT's architecture, training data, and overall hyper-parameter choices~\cite{devlin2019bert}. One noticeable difference, however, is that we follow RoBERTa to train the model without using a next-sentence prediction loss~\cite{liu2019roberta}, which has been shown to have a minimal impact on model performance. Each of the three models in our rerun not only has the exact same design choices in terms of training data, text processing, model architecture, etc., but is also implemented on the exact same codebase. Concretely, we use the BERT codebase as implemented on Huggingface, and conduct some slight modifications in terms of removing the next-sentence prediction loss as stated above. When implementing the other models, we take the BERT implementation, and simply change the masking function and pre-training objective in order to replicate the results of GPT-1 and ELMo under comparable conditions as our BERT model. This means that our reruns of the GPT-1 and ELMo models benefit from the exact same technical implementation details as our BERT model by virtue of using identical positional encoding, segment embeddings, etc. We tune the pre-training and fine-tuning learning rate of each model independently (hence the final learning rate for each model may be different), although we strive to dedicate the same amount of compute resources in tuning the hyper-parameters of each model, in order to avoid favoring one model over the others. We summarize some key design choices in Table~\ref{tab:exp-param}.

\subsection{Hyper-Paramaters for Fine-tuning}\label{app:fine-tuning-hyperparameters}
We refer to the original BERT's hyper-parameters for fine-tuning, but experiment with more fine-tuning learning rates. In addition to that, we also use three different random seeds. The fine-tuning hyper-parameters that we used are as follows, which is partially based on prior work \citep{joshi-etal-2020-spanbert}:
\begin{itemize}
  \itemsep0em
  \item \textbf{Batch sizes:} \{16, 32\}
  \item \textbf{Learning rates:} \{2e-4, 1e-4, 5e-5, 3e-5, 2e-5, 1e-5, 5e-6\}
  \item \textbf{Epoch:} 4
  \item \textbf{Random seeds}: \{1, 41, 386\}
\end{itemize}

\begin{table*}[!ht]\centering
\resizebox{0.99\textwidth}{!}{
\small
\begin{tabular}{cccccccc}
\toprule
Model           
    & Training Data
    & \begin{tabular}[c]{@{}c@{}}Text\\Processing\end{tabular}
    & \begin{tabular}[c]{@{}c@{}}Pre-Training\\Objective\end{tabular}      & Architecture 
    & \begin{tabular}[c]{@{}c@{}}Sequence\\Length\end{tabular}
    & \begin{tabular}[c]{@{}c@{}}Batch\\Size\end{tabular}
    & Steps/Epochs    \\
\midrule
BERT  Original         
    & \begin{tabular}[c]{@{}c@{}}English Wikipedia\\ Book Corpus\end{tabular} 
    & Wordpiece       
    & MLM, NSP
    & Transformers
    & 512                 
    & 256       
    & 1M steps \\
GPT-1 Original      
    & \begin{tabular}[c]{@{}c@{}}Word Benchmark\\ Book Corpus\end{tabular}    
    & BPE             
    & CLM     
    & Transformers
    & 512                
    & 64         
    & 100 epochs      \\
ELMO Original 
 & Word Benchmark 
 & \begin{tabular}[c]{@{}c@{}}Character level\\+ convolution \end{tabular}
 & CLM 
 & LSTM
 & 2048
 & N.A.
 & 10 epochs \\
\midrule
BERT Rerun      
    & \begin{tabular}[c]{@{}c@{}}English Wikipedia\\ Book Corpus\end{tabular} 
    & Wordpiece       
    & MLM   
    & Transformers
    & 512                 
    & 256        
    & 1M steps \\
GPT-1 Rerun     
    & \begin{tabular}[c]{@{}c@{}}English Wikipedia\\ Book Corpus\end{tabular} 
    & Wordpiece       
    & CLM
    & Transformers
    & 512                 
    & 256        
    & 1M steps \\
ELMO-variant Rerun 
    & \begin{tabular}[c]{@{}c@{}}English Wikipedia\\ Book Corpus\end{tabular} 
    & Wordpiece       
    & CLM          
    & Transformers
    & 512                 
    & 256        
    & 1M steps \\
\bottomrule
\end{tabular}
}
\caption{Comparison of the pre-training hyper-parameters across different models. MLM denotes masked language modelling; NSP denotes next-sentence prediction; CLM denotes causal language modelling, respectively. }
\label{tab:exp-param}
\end{table*}

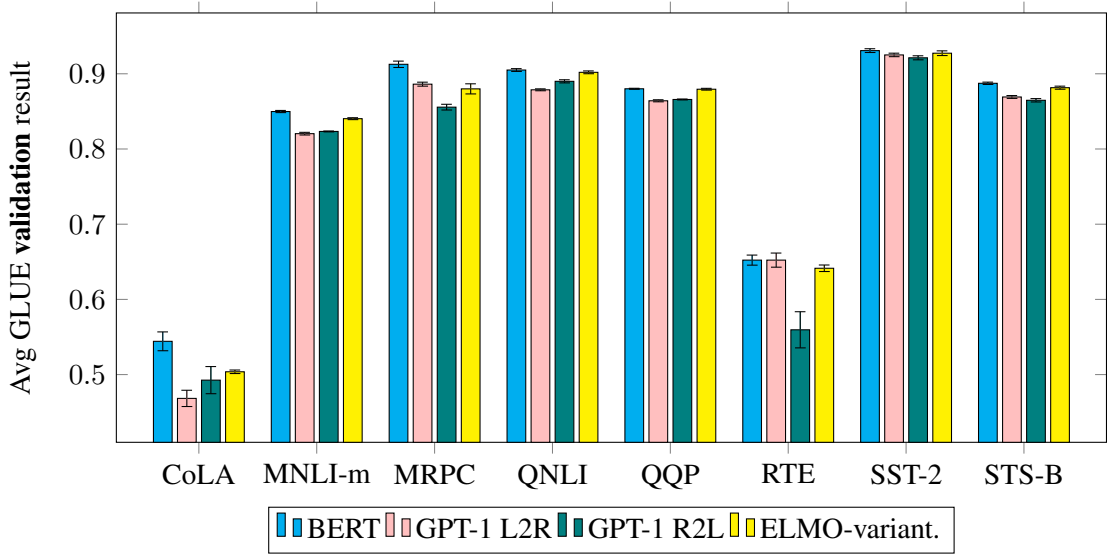
\begin{figure*}[ht!]
\centering
\begin{tikzpicture}
\begin{axis}[
    ybar,
    bar width=7,
    legend style={at={(0.5,-0.15)},
      anchor=north,legend columns=-1},
    ylabel={Avg GLUE \textbf{validation} result},
    symbolic x coords={CoLA,MNLI-m,MRPC,QNLI,QQP,RTE,SST-2,STS-B},
    xtick=data,
    x post scale=1.9,
]

\addplot [black, fill=cyan,]
 plot [error bars/.cd, y dir = both, y explicit]
 table[x=x, y=y, y error=ey]{BERTRerunv2.dat};
\addplot [black, fill=pink]
 plot [error bars/.cd, y dir = both, y explicit]
 table[x=x, y=y, y error=ey]{CausalL2Rv2.dat};
\addplot [black, fill=teal,]
 plot [error bars/.cd, y dir = both, y explicit]
 table[x=x, y=y, y error=ey]{CausalR2Lv2.dat};
\addplot [black, fill=yellow,]
 plot [error bars/.cd, y dir = both, y explicit]
 table[x=x, y=y, y error=ey]{ELMOv4v2.dat}; 
\legend{BERT, GPT-1 L2R, GPT-1 R2L,ELMO-variant.}
\end{axis}
\end{tikzpicture}
\caption{GLUE validation set results. For each task, we show the mean performance --- for the best validation set fine-tuning hyper-parameters as outlined in Appendix~\ref{app:fine-tuning-hyperparameters} --- alongside the standard deviation of the results (denoted with the error bars), which is computed based on three different random seeds for the exact same winning hyper-parameter. Note that the reported score for each task follows the exact same metric as the test result in Table~\ref{tab:test-1M}. The BERT, GPT-1, and ELMO-variant results reported here are based on our reruns with proper controls and comparable conditions between different models, and not from the original reported results of each work.  } \label{fig:appendix-c-v2}
\end{figure*}

\section{Extended Related Work}\label{app:related_work}
A number of prior work has made progress in disentangling the impact of different language modelling pre-training objectives by conducting principled, apple-to-apple comparisons under comparable experimental conditions; here we highlight four such prior work (among others), and remark on how our findings and recommendations complement theirs.

\citet{raffel2020exploring} introduced a unified, text-to-text format (\emph{i.e.,} encoder-decoder) framework, and conducted a systematic study over the impact of different pre-training objectives, architectures, and training datasets. More recently, \citet{tay2022unifying} conducted a series of comprehensive ablation experiments that compared the effectiveness of different pre-training objectives under comparable conditions. They found that interpolation of these objectives can be universally effective across different tasks, setups, and model scales.\footnote{A similar line of prior work \citep{unilm} also proposed combining multiple pre-training objectives using the same Transformer model through a Cloze-type formulation.} In line with our findings, their findings similarly demonstrate how conducting these systematic comparisons can lead to new insights and approaches that can rival or outperform current ones. Another line of work \citet{artetxe_2022} examined the role of \emph{bidirectionality} in language model pre-training. This is done by drawing a distinction between bidirectional context and bidirectional attention, generalizing how current approaches fall with respect to these spectrums, and characterizing the effects of each component on different downstream tasks. 

Our findings differ from --- and further complement --- this line of prior work in two ways. First, we revisit and augment the baseline ELMo model --- which incorporates a degree of bidirectionality at fine-tuning (albeit not at pre-training) --- with Transformer architectures and whole model fine-tuning, hence facilitating a fair comparison with the BERT model. We show that the resulting ELMo can achieve competitive performance with BERT in terms of overall GLUE performance; to our best knowledge, no such ELMo baseline was explored in prior work. We additionally demonstrate that a simple ensemble of left-to-right and right-to-left models, which are pre-trained and fine-tuned completely independently, can approximate the performance of BERT on the full data setup, and even \emph{outperform} it on the efficient learning scenario. Second, above and beyond conducting principled comparisons between BERT and its baselines, we outline several recommendations that would encourage and incentivize this line of work; we hope that these recommendations would be adopted by the broader community, with the aim of accelerating progress towards resolving our current scientific debt in language model pre-training research.

\section{Validation Set Results with Standard Deviation}\label{app:results_with_stdev}
In Figure~\ref{fig:appendix-c-v2}, we show the performance of our best validation set hyper-parameters for each task across three different random seeds, based on which we derive the error bars. We remark that the majority of the tasks have fairly small error bars, providing evidence for the robustness and generality of our observations across different random seeds.

\section{Dataset and Artifacts License}
\label{dataset-license}
The Wikipedia dataset is available under the Creative Common license, cc-by-sa-3.0, which we use solely for research purposes in accordance with its terms. 
 Our entire codebase is based on Huggingface’s open-source Transformer implementation, which is released under the Apache-2.0 license.

\end{document}